%% file: MainPaper.tex
\documentclass[10pt,journal,compsoc]{IEEEtran}

\ifCLASSOPTIONcompsoc
  \usepackage[nocompress]{cite}
\else
  \usepackage{cite}
\fi

\ifCLASSINFOpdf
  \usepackage[pdftex]{graphicx}
  \graphicspath{{../figures/}}
  \DeclareGraphicsExtensions{.pdf}
\else
  \usepackage[dvips]{graphicx}
  \graphicspath{{../figures/}}
  \DeclareGraphicsExtensions{.pdf}
\fi

\usepackage{amsmath}
\usepackage{array}
\usepackage{subfig}
\usepackage{booktabs}
\usepackage{algorithm}  
\usepackage{algorithmicx}
\usepackage{algpseudocode}  
\usepackage{multirow}
\usepackage{enumerate}
\usepackage{xcolor}
\usepackage{comment}
\usepackage{colortbl}
\usepackage{hyperref}

\usepackage[capitalize]{cleveref}
\crefname{section}{Sec.}{Secs.}
\Crefname{section}{Section}{Sections}
\Crefname{table}{Table}{Tables}
\crefname{table}{Tab.}{Tabs.}

\newcommand{\revise}[1]{\textcolor{black}{#1}}
\newcommand{\drevise}[1]{\textcolor{black}{#1}}

\begin{document}
\title{\revise{BAL: Balancing Diversity and Novelty \\ for Active Learning}}

\author{Jingyao~Li,
        Pengguang~Chen,
        Shaozuo~Yu,
        Shu~Liu,~\IEEEmembership{Member,~IEEE}
        and~Jiaya~Jia,~\IEEEmembership{Fellow,~IEEE}
\IEEEcompsocitemizethanks{\IEEEcompsocthanksitem Jingyao Li and Shaozuo Yu are with the Department of Computer Science and Engineering of the Chinese University of Hong Kong (CUHK) \\
Jingyao Li's E-mail: jingyao.li@link.cuhk.edu.hk
\IEEEcompsocthanksitem Pengguang Chen, Shu Liu, and Jiaya Jia are with SmartMore.}
\thanks{Manuscript received Feb 8th, 2023.}}


\IEEEtitleabstractindextext{%
\input{sections/01_abstract.tex}
}

\maketitle
\IEEEdisplaynontitleabstractindextext
\IEEEpeerreviewmaketitle

\IEEEraisesectionheading{\section{Introduction}\label{sec:intro}}
\input{sections/02_intro.tex}

\section{Related Works}\label{sec:related}
\input{sections/03_relatex_works.tex}

\section{Methods}\label{sec:methods}
\input{sections/04_methods.tex}

\section{Experiments}\label{sec:exp}
\input{sections/05_experiment.tex}

\section{Ablation Experiments}\label{sec:ablation}
\input{sections/06_ablation.tex}

\section{\revise{Discussion}}\label{sec:limit}
\input{sections/07_limitation.tex}

\section{Conclusion}\label{sec:conclusion}
\input{sections/08_conclusion.tex}

\ifCLASSOPTIONcompsoc
\else
\fi


\ifCLASSOPTIONcaptionsoff
  \newpage
\fi

\bibliographystyle{IEEEtran}
\bibliography{egbib}

\appendices
\input{sections/09_bio.tex}

\end{document}

%% file: sections/01_abstract.tex
\begin{abstract}
\drevise{The objective of Active Learning is to strategically label a subset of the dataset to maximize performance within a predetermined labeling budget. In this study, we harness features acquired through self-supervised learning. We introduce a straightforward yet potent metric, Cluster Distance Difference, to identify diverse data. Subsequently, we introduce a novel framework, Balancing Active Learning (BAL), which constructs adaptive sub-pools to balance diverse and uncertain data. Our approach outperforms all established active learning methods on widely recognized benchmarks by 1.20\%. Moreover, we assess the efficacy of our proposed framework under extended settings, encompassing both larger and smaller labeling budgets. Experimental results demonstrate that, when labeling 80\% of the samples, the performance of the current SOTA method declines by 0.74\%, whereas our proposed BAL achieves performance comparable to the full dataset. Codes are available at \href{https://github.com/JulietLJY/BAL}{https://github.com/JulietLJY/BAL}.}
\end{abstract}

\begin{IEEEkeywords}
Active Learning, Computer Vision, Contrastive Learning
\end{IEEEkeywords}

%% file: sections/02_intro.tex
\IEEEPARstart{R}educing \drevise{the time and cost associated with data labeling has posed a longstanding challenge for deploying large-scale deep models across various computer vision tasks \cite{imagenet, resnet, deeplab,fcn, tagclip, mood}. Several algorithms have been developed to address labeling costs, including semi-supervised learning \cite{semi-supervised_learning}, weakly supervised learning \cite{weakly-supervised_learning}, few-shot learning \cite{few-shot_learning}, and active learning \cite{active_learning}. Among these, Active Learning (AL) stands out as a facilitator, enabling the selection of the most valuable data to achieve optimal performance within a fixed labeling budget. The AL process commences with an unlabeled pool of samples. In each cycle, $K$ additional samples, equivalent to the budget, are selected for labeling. Previous studies have primarily focused on small sample budgets, inevitably resulting in diminished model performance compared to results obtained with the complete dataset. However, our proposed active learning approach reveals that judiciously selecting a subset of data can yield comparable results to using the entire dataset. Therefore, our method proves valuable for conserving labeling budgets in scenarios where performance decline is not acceptable.}

\drevise{The conventional approach in active learning involves selecting diverse or uncertain samples. Distribution-based methods sample data from high-density regions \cite{settles2008analysis, settles2008curious} that effectively represent the overall feature distribution. Conversely, uncertainty-based methods \cite{lewis1994sequential} concentrate on sampling the most uncertain data, often measured by posterior probabilities \cite{lewis1994heterogeneous, lewis1994sequential}, entropy \cite{joshi2009multi, shannon2001mathematical}, among other metrics. Earlier active learning methods \cite{settles2008analysis, lewis1994heterogeneous} traditionally relied on the primary task network to select diverse or uncertain samples. Subsequent studies \cite{PAL, PT4AL}, however, employ pretext tasks as scoring networks. For acquiring diverse data, pretext-based methods create multiple sub-pools from the entire dataset based on specific properties. In each cycle, they employ uncertainty-based samplers to select from the corresponding sub-pool, resulting in labeled samples within each sub-pool. Consequently, these samples collectively represent the entire data distribution, while uncertainty-based samplers ensure diversity. The design of the indicator for generating sub-pools is crucial. Existing pretext-based methods typically leverage the output of an additional self-supervised head \cite{PAL} or the pretext task loss \cite{PT4AL}. However, we observe that neither an additional head nor pretext loss achieves comparable performance with the finely learned features of the pretext task (\cref{sec:exp}). These features encompass information on inter-sample relationships and significantly benefit downstream tasks.}

\begin{figure*}[!t]
\centering
\subfloat[Self-Supervised Task]{\includegraphics[width=\linewidth,trim=10 380 30 30, clip]{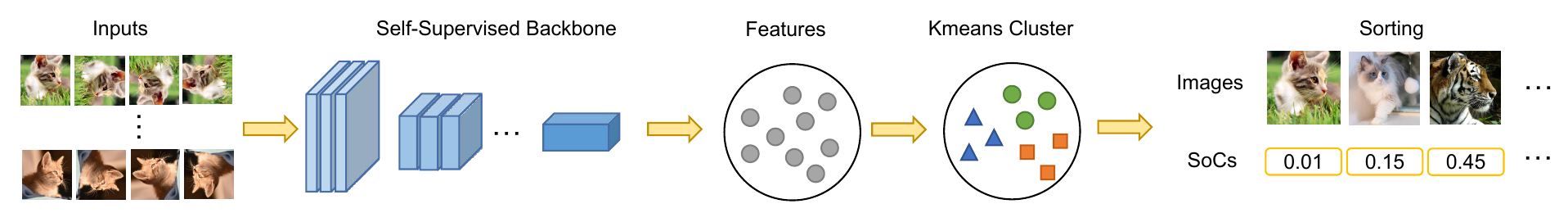}%
\label{fig:BAL_SS}}
\hfill
\subfloat[Active Learning Module]{\includegraphics[width=\linewidth,trim=10 180 30 180,clip]{figures/BAL.pdf}%
\label{fig:BAL_AL}}
\caption{\drevise{Our Balancing Active Learning (BAL) framework consists primarily of a self-supervised task and an active learning module. \textbf{(a) Self-Supervised Task}: In this phase, we train a self-supervised model on the unlabeled pool. Subsequently, we apply K-means clustering to the features and calculate their Cluster Distance Differences (CDDs) to sort the data. \textbf{(b) Active Learning Module}: In this phase, we generate sub-pools from the sorted unlabeled pool and employ an uncertainty-based sampler to select data from the corresponding sub-pool in each cycle.}}
\label{fig:BAL}
\end{figure*}

\drevise{In this paper, we initially cluster features derived from the self-supervised pretext task. We then introduce a straightforward yet impactful indicator known as Cluster Distance Difference (CDD) within the feature space. CDD attains its minimum point, zero, when a feature resides on the decision boundary of the two nearest cluster centers, signifying the most challenging sample to differentiate. Our approach substitutes the decision surface solution with a distance difference, achieving computational efficiency while maintaining the desired effectiveness. CDD surpasses the previous indicator, the pretext task loss, in the current state-of-the-art method, PT4AL \cite{PT4AL}, by 1.36\%. Additional results are detailed in \cref{sec:abcdd}. Another critical aspect is balancing uncertainty and diversity to label a more reasonable distribution of samples. Existing methods, such as PT4AL \cite{PT4AL}, employ equidistant disjoint sub-pools, but determining the size of each sub-pool poses challenges. A small sub-pool has a greater impact on the order of CDDs, limiting the choice of the uncertainty-based sampler. For instance, when the sub-pool contains only $K$ samples, the selected data comprises the entire sub-pool, rendering the uncertainty-based sampler unused. Conversely, with a large sub-pool, selected samples exhibit more uncertainty but less diversity. In extreme cases where sub-pools encompass the entire dataset in cycles, the pretext-based approach regresses into a simplistic uncertainty-based method.}

\drevise{In our study, we introduce the Balancing Active Learning (BAL) framework, as illustrated in \cref{fig:BAL}, to tackle this challenge. Unlike conventional pretext-based approaches, BAL facilitates overlaps between adjacent sub-pools. It dynamically modifies the size of each sub-pool based on early performance using a small labeled pool. Consequently, the number of samples in each sub-pool can dynamically align with the demands of various labeling budgets. Further insights can be found in \cref{sec:al}.} 

\drevise{In our experimental evaluation, our approach demonstrates a superior performance compared to the current SOTA across multiple datasets, achieving a notable improvement of 1.20\% on widely recognized benchmarks, as detailed in \cref{sec:exp_small}. Furthermore, we assess our active learning approach under large labeling budgets in \cref{sec:exp_large}. Notably, when labeling 80\% of samples, the performance of the existing SOTA method \cite{PT4AL} diminishes by 0.74\%, while our BAL approach maintains its efficacy. These results suggest that our sampled data provides comparable information to represent the entire dataset. Subsequently, we investigate the performance of BAL under small labeling budgets. Notably, BAL consistently outperforms typical methods by substantial margins, even outperforming current SOTAs in common benchmarks, such as PT4AL~\cite{PT4AL} and ActiveFT~\cite{activeft}. Particularly, when training on only 4\% of the total dataset, BAL exhibits an impressive performance advantage, surpassing the current SOTA by 11\% on CIFAR-10 and by 26\% on SVHN.}


In summary, our main contributions are threefold:
\drevise{
\begin{enumerate}
\item We introduce the Cluster Distance Difference (CDD) as a simplified yet effective indicator for measuring the distance to the decision surface, providing a nuanced measure of clustering difficulty.
\item We propose the Balancing Active Learning (BAL) framework, which adeptly balances diversity and uncertainty, enabling the adaptive selection of optimal data across various settings.
\item We demonstrate BAL's superior performance over existing active learning methods on diverse datasets and labeling budgets, establishing its robustness and versatility.
\end{enumerate}
}

%% file: sections/03_relatex_works.tex
\subsection{Active learning}
\label{sec:related_al}
\drevise{The active learning strategy typically involves sampling data with substantial uncertainty or diversity.}

\drevise{\textbf{Uncertainty} (e.g., novelty, confusion) in active learning refers to a sample's ability to provide new information independently of other labeled samples. Various uncertainty methods employ different metrics to measure uncertainty, including predicted class posterior probability \cite{lewis1994sequential, lewis1994heterogeneous}, the difference between the first and second class predicted posterior probabilities \cite{joshi2009multi, roth2006margin}, entropy of samples \cite{luo2013latent, joshi2009multi}, distance from the support vector machine to the decision boundary \cite{tong2001support, vijayanarasimhan2014large, li2014multi}, inconsistencies between a committee of multiple independent models \cite{seung1992query, mccallumzy1998employing, iglesias2011combining}, Bayesian frameworks \cite{gal2016dropout, DBAL}, etc.}

\drevise{\textbf{Diversity} (e.g., representativeness, coverage) refers to a sample's ability to represent the distribution of the unlabeled data effectively. In existing uncertainty methods, information density approaches \cite{settles2008analysis, settles2008curious} weigh the informativeness of a sample by its similarity to other data in the input distribution. The nearest-neighbor method \cite{fujii1999selective} selects samples that are most unlike the labeled instances and similar to the unlabeled samples. Coreset \cite{coreset} is a method based on identifying a core set that models the empirical loss over the set of labeled samples and the pool of query samples. VAAL \cite{vaal} is designed to learn a good representation using a variational autoencoder.}

\drevise{The drawback of distribution-based approaches is that data near the classification boundary may confuse the model. In contrast, uncertainty-based approaches may sample overlapping data and find it challenging to extract a representation of the entire data distribution. In comparison, our work leverages self-supervised features to balance diverse and uncertain samples, benefiting from both aspects.}


\subsection{Representational Learning}
\label{sec:related_rl}
\drevise{In representational learning, self-supervised models trained on unlabeled datasets acquire features transferable to downstream tasks. Common self-supervised transformations include color removal \cite{liu2021influence}, resolution reduction \cite{fcn}, partial image obscuring \cite{pathak2016context}, spatial order confusion of sub-images \cite{mccallumzy1998employing}, and random geometric transformations \cite{vgg}, among others. Features learned through these tasks are subsequently applied to more complex downstream tasks \cite{imagenet, resnet}. Recent work \cite{simclr, simsiam, moco} showcases advanced results and underscores the robustness of contrastive learning in various downstream tasks.}

\drevise{The latest active learning approaches \cite{PAL, PT4AL} incorporate self-supervised learning as a pretext task, yielding impressive results. PAL \cite{PAL} relies on the output of an additional self-supervised head trained in parallel with the task network. However, the self-supervised head is inadequately trained, limiting its performance. PT4AL \cite{PT4AL} pre-trains the self-supervised task network first and employs its losses to partition unlabeled samples for cycles. Nevertheless, pretext task losses are generally designed to update parameters rather than facilitate downstream tasks.}

\drevise{In contrast to prior pretext-based methods, BAL (i) harnesses finely-learned self-supervised features that encompass information on inter-sample relationships, benefiting downstream tasks significantly, and (ii) dynamically adjusts the length of adaptive sub-pools to meet the requirements of different labeling budgets. In \cref{sec:exp_small}, experiments validate that BAL outperforms all previous active learning methods across multiple datasets.}

%% file: sections/04_methods.tex
In this section, we introduce our proposed Balancing Active Learning (BAL) framework. Its structure is shown in \cref{fig:BAL}.

We first define the notations. A typical active learning scenario consists of the unlabeled data pool $X_U\in X$ and labeled data $X_L\in X$, where $X$ is the dataset. The goal of an active learner is to learn an effective model $F_m(\cdot)$ with limited labeling budgets. In the $i$-th cycle, samples are selected from $X^i_U$ and labeled by the active learner, resulting in newly labeled data $X^i_K$. The labeled pool is updated according to $X^i_L=X^{i-1}_L\cup X^i_K$ and main task model $F^i_m(\cdot)$ is trained on a new labeled set $X^i_L$.

\subsection{Balancing Active Learning}
Active learning approaches usually select uncertain \cite{lewis1994sequential} or diverse \cite{settles2008analysis} data. Uncertain data represents the data close to decision boundaries, while diverse data defines distribution in the feature space well. Existing research \cite{coreset, PT4AL, PAL} with a simple combination of the two aspects hardly reach the balance, which varies according to different proportion of labeling. To extract a more reasonable distribution of samples, we propose Balancing Active Learning (BAL). Its main process is as follows.
\begin{enumerate}
\item Train a self-supervised model $F_{ss}$ on unlabeled $X$.
\item Cluster features $\{f_j\}_{j=1}^N$ extracted from $F_{ss}$ and sort $X$ in ascending CDD order.
\item Automatically set balancing factor $\beta$  based
on the maximal performance of the main task model $F_m$ trained on the early labeled pool.
\item Generate adaptive sub-pools $\{P_U^i\}^I_{i=0}$ from $X$.
\item Select $K$ samples from $P_U^i$ for labeling based on posterior probabilities at $i$-th cycle.
\item Repeat Step 4 until meeting labeling budgets.
\end{enumerate}

In each cycle, we sample from the corresponding sub-pool. Thus there exist labeled samples in each sub-pool. Consequently, these samples together represent the whole data distribution. On the other hand, the uncertainty-based sampler ensures novelty. Our adaptive sub-pools balance the two aspects in \cref{sec:al}.

In the following sections, we introduce two stages of our BAL framework, including the self-supervised task in \cref{sec:ss} and the active learning module in \cref{sec:al}. 

\subsection{Self-Supervised Task}
\label{sec:ss}

\noindent\textbf{Training Strategy.}
The main purpose of the self-supervised task module is to extract features that can be well passed on to the downstream active learning task. Past researchers \cite{PAL,PT4AL} have found that compared to the latest self-supervised learning methods such as SimCLR \cite{simclr} and SimSiam \cite{simsiam}, rotation prediction \cite{gidaris2018unsupervised} is better at self-supervised task learning. The loss function is defined as
\begin{equation} \label{equ:L_SS}
L_{ss}(x_i)=\frac{1}{k}\sum_{\theta\in\{0,90,180,270\}}L_{CE}(l_{ss}(t(x_i, \theta)), y_{\theta}),
\end{equation} 
where $L_{CE}$ is the cross-entropy loss. $t(x_i, \theta)$ represents an image that rotates the input image $x_i$ by $\theta$ degree. $l_{ss}(\cdot)$ represents the output of the self-supervised model $F_{ss}(\cdot)$. $y_{\theta}$ represents labels.

\begin{figure}[t]
\centering
\subfloat[Cluster distance difference]{\includegraphics[width=0.49\linewidth,trim=0 0 480 0,clip]{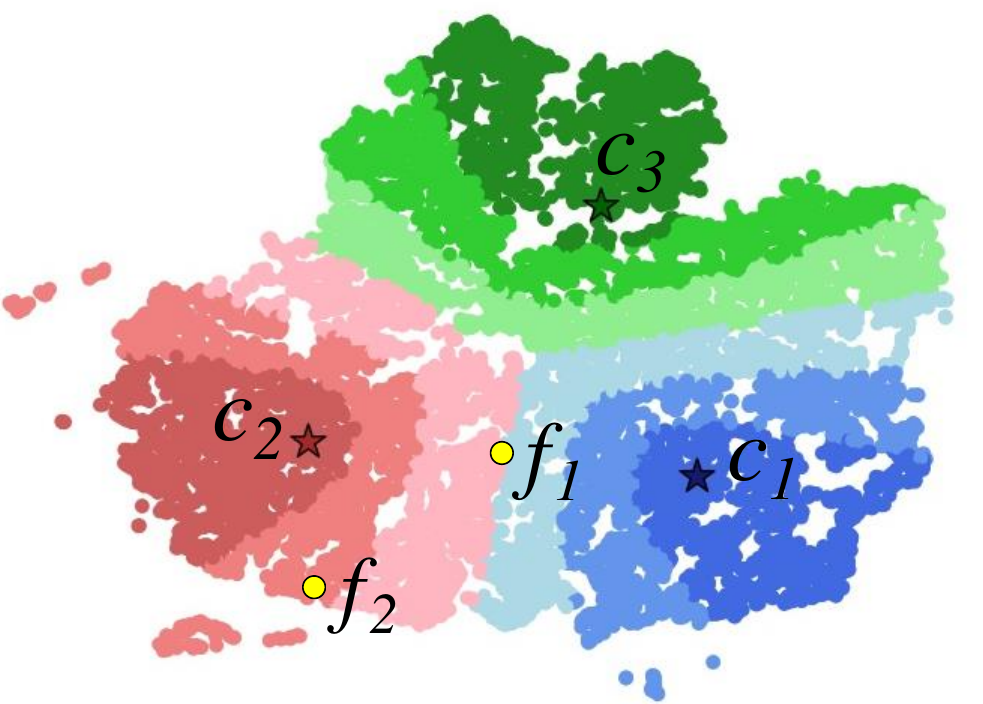}}
\subfloat[Distance to the nearest center]{\includegraphics[width=0.49\linewidth,trim=480 0 0 0,clip]{figures/CDD.pdf}}
\caption{Visualization of three K-means clusters of self-supervised features on CIFAR-10. Stars represent cluster centers {\color{blue}$c_1$}, {\color{red}$c_2$}, and {\color{teal}$c_3$}. The three colors represent three clusters. Different shades of colors represent different sub-pools divided according to (a) Cluster distance difference and (b) Distance to the nearest center.}
\label{fig:CDD}
\end{figure}

\subsubsection{Cluster Distance Difference.}
After training of self-supervised model $F_{ss}$, we leverage its features $\{f_i\}_{i=1}^N$ to sort the unlabeled dataset $X$, as shown in \cref{fig:BAL_SS}. It has been experimentally verified that the performance of the pretext task and that of the main task has a strong positive correlation \cite{PT4AL}, which supports our utilization of self-supervised features. 

Firstly, we perform K-means clustering \cite{kmeans} on $\{f_i\}_{i=1}^N$ to cluster them into $N_k$ clusters. Next, we design $\alpha (f_i)$ as the difference between $f_i$ to the two nearest cluster centers, to measure the difficulty of clustering $f_i$.
\begin{equation}
\begin{aligned}
d_1 (f_i)&=\min_{1\le j\le N_k}\{||f_i, c_j||_2^2\}, \\
d_2 (f_i)&=\min_{1\le j\le N_k, j\ne \arg\min_j\{||f_i, c_j||_2^2\}}\{||f_i,c_j||_2^2\}, \\
\alpha (f_i) &= d_2 (f_i) - d_1 (f_i), \\
\end{aligned} \label{equ:CDD}
\end{equation}
where $||f_i, c_j||_2^2$ is Euclidean distance from $f_i$ to cluster center $c_j$. $d_1 (f_i)$ represents distance from $f_i$ to the nearest cluster center and $d_2(f_i)$ represents the second shortest distance. $\alpha(f_i)$ is the value of our proposed indicator, cluster distance difference (CDD), of $f_i$. 

Finally, we sort $X$ in ascending CDD order, resulting in sorted data $X_S$, which starts from samples with the smallest CDD, i.e., the most difficult sample to distinguish.

\revise{
\subsubsection{Explanation of Cluster Distance Difference}  
In a simplified scenario, we ignore certain situations such as non-convex decision boundaries or overlap between clusters. Our goal is to find a dividing hyperplane (threshold) of the two clusters, which is defined by the equality of the distances from data point $f_i$ to the two nearest centers $c_1$ and $c_2$:
\begin{equation}\label{equ:d1=d2}
    d_1 (f_i) = d_2 (f_i).
\end{equation}
}
\revise{
Among \cref{equ:d1=d2}, if $d_1 (f_i)$ is smaller, $f_i$ is closer to Cluster 1; if $d_2 (f_i)$ is smaller, $f_i$ is closer to Cluster 2. When \cref{equ:d1=d2} holds true, $f_i$ lies on the boundary hyperplane, indicating that it is situated on the border between the two clusters. Therefore, \cref{equ:d1=d2} signifies that the boundary hyperplane can be represented as:
\begin{equation}\label{equ:a=0}
    \alpha (f_i) = 0,
\end{equation}
}
\revise{where $f_i$ represents the data point in the feature space and $\alpha$ is its CDD.} When the CDD $\alpha(f_i)$ is smaller, it is more difficult to distinguish which cluster $f_i$ belongs. For example, when applying the CDD to the scenario shown in \cref{fig:CDD}, 
\begin{equation}\label{equ:CDD}
\begin{aligned}
\alpha (f_1)&=||f_1, c_1||_2^2-||f_1, c_2||_2^2=0 \\ &<||f_2, c_1||_2^2-||f_2, c_2||_2^2\\&=\alpha (f_2),
\end{aligned}
\end{equation}
which represents that $f_1$ is more difficult to be distinguished than $f_2$. When a feature $f_i$ is located on the decision plane, $\alpha(f_i)=0$ reaches the lowest point, representing the most difficult to distinguish.

Our method replaces the solution of the decision surface with a distance difference, which saves large computation cost as well as achieve the desired effect. In comparison, when applying a naive baseline, distance to the nearest cluster center, in \cref{fig:CDD}, it mistakenly regards $f_2$ as more difficult to cluster. 

When comparing with another indicator, loss of pretext task, in the current SOTA \cite{PT4AL}, CDD outperforms it by 1.36\% (\cref{sec:ablation}). We explain that the pretext task loss is designed to update the parameters instead of passing them on to downstream tasks. On the contrary, well-learned self-supervised features include the information of inter-sample relationships and benefit downstream tasks outstandingly.

\subsection{Active Learning Module}
The algorithm of our Balancing Active Learning framework is shown in \cref{alg1}.
\label{sec:al}

 \subsubsection{Adaptive Sub-pools.}
We generate sub-pools $\{P_U^i\}_{i=1}^I$ from $X_S$ as
\begin{equation} \label{equ:P_U}
P_U^i(\beta) = 
\begin{cases}
  & \{x_k\}_{k=1}^{\beta\frac{N}{I}}, \text{if }i=1, \\
  & \{x_k\}_{(k=i+\frac{1-\beta}{2})\frac{N}{I}}^{(i+\frac{\beta+1}{2})\frac{N}{I}} / \{X_L^j\}_{j=1}^{i-1}, \text{if }1<i<I, \\
  & \{x_k\}_{k=N-\beta\frac{N}{I}}^{N} / \{X_L^j\}_{j=1}^{i-1},\text{if }i=I, \\
\end{cases}
\end{equation}
where $X_L^i$ is the labeled pool in the $i$-th cycle. $I$ is the number of cycles. $N$ is the number of total images. $x_k$ is the $k$-th sample in the sorted dataset $X_S$, and $\beta$ is the balancing factor of diversity and uncertainty. The number of samples in each $P_U^i$ is as
\begin{equation} \label{equ:|P_U|}
|P_U^i| = \left \lfloor \beta \frac{N}{I}\right \rfloor. 
\end{equation}

\begin{algorithm}[t]
\renewcommand{\algorithmicrequire}{\textbf{Require:}}
\caption{Balancing Active Learning Algorithm}
\label{alg1}
\begin{algorithmic}[1]

\Require Unlabeled dataset $X$ with $N$ samples, untrained main task model $F^0_m$, self-supervised task model $F_{ss}$, number of cycles $I$, balancing factor candidates $\{\beta_j\}_{j=1}^{N_\beta}$
\Ensure Trained main task model $\{F^i_m\}_{i=1}^{I}$.
\State Train $F_{ss}$ with the loss $L_{ss}$ in \cref{equ:L_SS} on unlabeled $X$.
\State K-means Cluster features $\{f_i\}_{i=1}^N$ extracted from $F_{ss}$.
\For{$i=1$ to $I$} 
\State Compute CDDs $\alpha(f_i)$ by \cref{equ:CDD}.
\EndFor
\State Sort the unlabeled dataset $X$ in ascending CDD order to get the sorted dataset $X_S=\{x_k\}_{k=1}^N$.
\State $X_L^1 = \{x_k\}_{k=1}^K$, where $x_k$ is the $k$-th sample in $X_S$. 
\State Train $F^1_m$ on $X_L^1$.
\For {$j=1$ to $N_\beta$} 
\State Label $\psi(P_U^{2}(\beta_j))$ , where $P^2_U$ and $\psi$ are defined by \cref{equ:P_U,equ:X_K}.
\State Train $F_m^{2}(\beta_j)$ on $X_L^1\cup \psi(P_U^{2}(\beta_j))$.
\EndFor
\State Compute $\beta$ according to \cref{equ:beta}.
\For{$i=2$ to $I$} {\color{gray} \Comment{Active learning.}}
    \State Label $X_K^i$, which is defined by \cref{equ:X_K}.
    \State Train $F^i_m$ on $X_L^i=X_L^{i-1}\cup X_K^i$.
\EndFor
\end{algorithmic}
\end{algorithm}

The minimum value of $\beta$ is $\left \lceil\frac{KI}{N} \right \rceil $, where $K$ is the number of samples to label in each cycle. It is because that $P_U^i$ requires at least $K$ samples for labeling.

\subsubsection{Sampling Method.}
In the first cycle, the sampler $\psi$ selects the first $K$ data from sorted unlabeled pool $X_S$. To obtain uncertain data, in the $i$-th ($i>1$) cycle, $\psi$ selects $K$ data with the lowest maximum posterior probability \cite{lewis1994sequential,PT4AL}, calculated by $F^{i-1}_m$ from  $P_U^i$. Thus, the newly labeled data $X_K^i$ is as
\begin{equation} \label{equ:X_K}
\begin{aligned}
X_K^i &= \psi(P^i_U) \\ &=
\begin{cases}
& \{x_k\}_{k=1}^K, \text{if } i=1, \\
& \arg\min_K\{\max\{l^{i-1}_m(P_U^{i})\}\}, \text{if } i>1,\\
\end{cases}
\end{aligned}
\end{equation}
where $l_m^{i-1}$ is output of main task model $F_m^{i-1}$ trained in the $(i-1)$-th cycle. $P_U^i$ is the $i$-th sub-pool, and $K$ is the number of data to label in each cycle. 

\begin{table*}[t]
\centering
\small
\setlength{\tabcolsep}{3.7mm}
\begin{tabular}{c|c|cccccccc}
\toprule
Dataset& Stage& Epochs & Momentum & Learning Rate & Batch Size & Scheduler  & Weight Decay     \\
\midrule
\multirow{2}{*}{CIFAR10} & Pretext & 120 & 30,60,90 & 0.01  & 256  & multi-step & $5\times10^{-4}$ \\
 & Main & 100 & 30,60,90 & 0.1& 128  & multi-step  & $5\times10^{-4}$ \\
\midrule
\multirow{2}{*}{SVHN} & Pretext & 120 & 30,60,90 & 0.01  & 256  & multi-step & $5\times10^{-4}$ \\
 & Main & 100 & 80 & 0.1& 128  & multi-step  & $5\times10^{-4}$ \\
\midrule
\multirow{2}{*}{Caltech101} & Pretext & 120 & 30,60,90 & 0.001 & 16& multi-step  & $5\times10^{-4}$ \\
 & Main & 100 & 80 & 0.001 & 16& multi-step  & $5\times10^{-4}$\\
\bottomrule
\end{tabular}
\caption{Experimental Configuration.}
\label{tab:cfgs}
\end{table*}

\subsubsection{Balancing Factor.} Existing approaches \cite{PAL, PT4AL} only utilize equidistant disjoint sub-pools, which cannot meet varying requirements. When $|P_U^i|$ is small, how to derive sub-pools $\{P_U^i\}_{i=1}^I$ has greater influence, while choices of uncertainty-based sampler $\psi$ is constrained. For example, when $\beta=\left \lceil\frac{KI}{N} \right \rceil$, there are only $K$ samples $P_U^i$ ($|P_U^i|=K$). Then uncertainty-based sampler $\psi$ has no choice but label the total sub-pool $P_U^i$, that is, $X_U^i=P_U^i$. It means that $\psi$ is unutilized. On the contrary, when $|P_U^i|$ is large, samples are mostly gathered along the decision boundary and become less diverse. In the extreme case that $P_U^i$ is the complete dataset $X_U$, the pretext-based approach deteriorates to a simple uncertainty-based method. 

In our work, the balance of uncertainty and diversity is reached by a balancing factor $\beta$. Our BAL allows adjacent sub-pools $P_U^i$ to have overlaps ($\beta>1$) or intervals ($\beta>1$), as shown in \cref{fig:BAL_AL}. Thus the length of each sub-pool $|P_U^i|$ can dynamically meet the requirements of different labeling budgets. The balancing factor $\beta$ is chosen based on the maximal performance of the main task model $F_m$ trained on $X_L^2$, as
\begin{equation} \label{equ:beta}
\beta = \beta_{j^*} \text{, where } j^* = \arg \mathop{\max}_{1\le j \le N_\beta}{\{F^{2}_m(\beta_j)\}},
\end{equation}
where $F^{2}_m(\beta_j)$ is trained on $X_L^1\cup \psi(P_U^{2}(\beta_j))$, which is quite small (just 15\% in common experiments). Thus, the balance of diversity and uncertainty can be easily reached without much computation requirements. 

In \cref{sec:exp_small}, experiments show that our adaptive sub-pools outperform uniformly-divided sub-pools by 1.0\% on Caltech-101. The algorithm of Balancing Active Learning is in \cref{alg1}.

\subsubsection{Cases Analysis.} Next, we analyze different cases of the balancing factor $\beta$. From \cref{equ:P_U}, when $2<i<I$, $\{P_U^j\}_{j=1}^{i-1}$ ends at the $(i+\frac{\beta-1}{2})$-th sample while $P_U^{i}$ starts at the $(i+\frac{1-\beta}{2})$-th sample. 

\vspace{3mm}\noindent\emph{Case1: $\beta > 1$.} In this case, we have $i+\frac{1-\beta}{2}>i+\frac{\beta-1}{2}$, so there exists an overlap of $P_U^i$ and $\{P_L^j\}_{j=1}^{i-1}$. Since some samples may have already been labeled in previous cycles, we need to remove $\{X_L^j\}_{j=1}^{i-1}$ from $P_U^i$, as shown in \cref{equ:P_U}. 


\vspace{3mm}\noindent\emph{Case2: $\beta < 1$.} In this case, we have $i+\frac{1-\beta}{2} < i+\frac{\beta-1}{2}$, and there exists an interval of $P_U^i$ and $\{P_L^j\}_{j=1}^{i-1}$. Thus, \cref{equ:P_U} can be simplified as
\begin{equation} \label{equ:beta<1}
P_U^i(\beta) = 
\begin{cases}
  & \{x_k\}_{k=1}^{\beta\frac{N}{I}}, \text{if }i=1, \\
  & \{x_k\}_{k=(i+\frac{1-\beta}{2})\frac{N}{I}}^{(i+\frac{\beta+1}{2})\frac{N}{I}}, \text{if }1<i<I, \\
  & \{x_k\}_{k=N-\beta\frac{N}{I}}^{N},\text{if }i=I. \\
\end{cases}
\end{equation}

\vspace{3mm}\noindent\emph{Case3: $\beta = 1$.} In this case, we have $|P_U^i|=\left \lfloor\frac{N}{I} \right\rfloor$, and there is no overlap or interval between nearby sub-pools, which shows that we obtain the sub-pools $\{P_U^i\}_{i=1}^I$ by uniformly divide the sorted dataset $X_S$. Therefore, \cref{equ:P_U} deteriorates to evenly-divided sub-pools as
\begin{equation} \label{equ:beta=1}
P_U^i(1) = \{x_k\}_{k=i\frac{N}{I}}^{(i+1)\frac{N}{I}}, 1\le i\le I, 
\end{equation}
where $I$ is the number of cycles. $N$ is the number of images and $x_k$ is the $k$-th sample in the sorted dataset $X_S$.




%% file: sections/05_experiment.tex
\begin{figure*}[t]
\centering
\subfloat[CIFAR-10]{\includegraphics[width=0.25\linewidth,trim=0 0 0 0,clip]{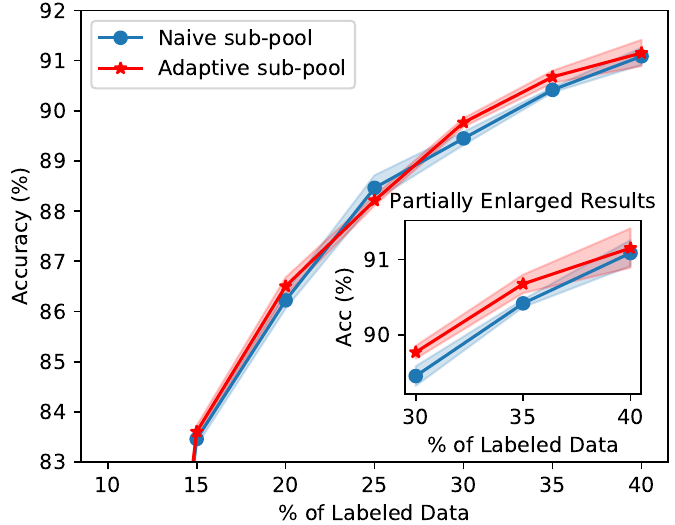}}
\subfloat[SVHN]{\includegraphics[width=0.25\linewidth,trim=0 0 0 0,clip]{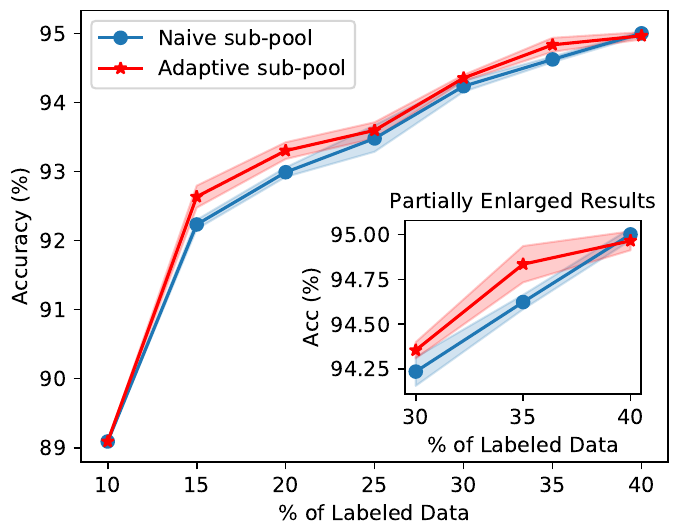}}
\subfloat[Caltech-101]{\includegraphics[width=0.25\linewidth,trim=0 0 0 0,clip]{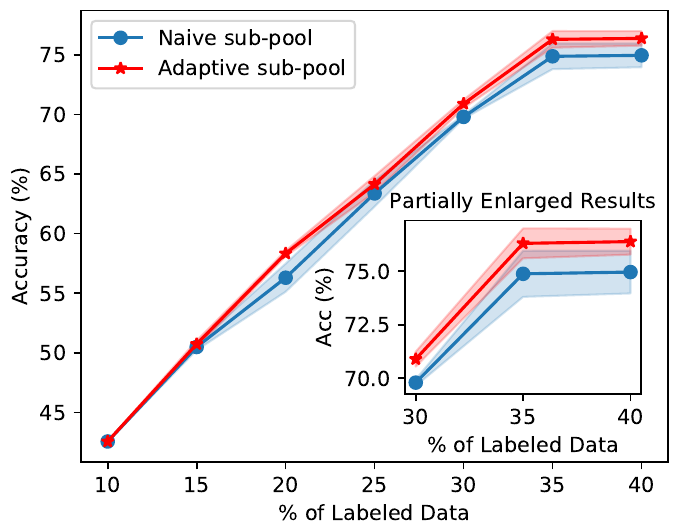}}
\subfloat[\revise{Tiny-ImageNet}]{\includegraphics[width=0.25\linewidth,trim=0 0 0 0,clip]{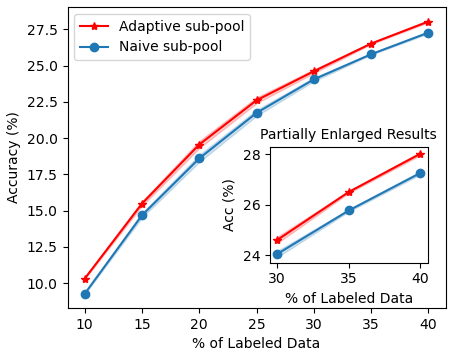}}
\caption{Comparison between our adaptive sub-pools and naive sub-pools on (a) CIFAR-10, (b) SVHN, (c) Caltech-101 \revise{and (d) Tiny-ImageNet}. Both of the sub-pools utilize CDD as sorting metrics. Light colors represent the variance. Small figures show partially enlarged results.}
\label{fig:beta}
\end{figure*}

In this section, we perform various experiments for different labeling budgets to demonstrate the effectiveness of our proposed Balancing Active Learning (BAL) framework. In \cref{sec:exp_config}, we introduce included datasets, competitive approaches, and implementation details. In \cref{sec:exp_med}, we perform our approach for medium labeling budgets and compare it with the most advanced methods. In \cref{sec:exp_small}, we test our method for small labeling budgets to further demonstrate its robustness. In \cref{sec:exp_large}, experiments with large data sampling are performed and it has been shown that our proposed method on carefully selected 80\% of the dataset could achieve the result on the full dataset. In \cref{sec:vis}, we illustrate some visualization for vivid exhibition.
 
\subsection{Configuration}
\label{sec:exp_config}
In this section, we introduce our experimental configuration details including datasets, competitive approaches, configurations and labeling budgets.

\subsubsection{Datasets.} 
Our experiments are conducted on extensive datasets including:
\begin{enumerate}
    \item SVHN \cite{svhn}, from Google Street View images, 10 categories of $32\times32$-pixel images.
    \item CIFAR-10 \cite{cifar10}, 10 categories, each containing 6000 $32\times32$-pixel images.
    \item Caltech-101 \cite{caltech101}, 101 uneven categories, each containing between 40 and 800 $300\times200$-pixel images.
    \item \revise{Tiny-ImageNet \cite{imagenet}, 200 classes of $256\times256$-pixel images.}
\end{enumerate}

\subsubsection{Competitive approaches.} We compared our approach to the following active learning strategies: 
\begin{enumerate}
    \item Random sampling: which is the simplest baseline.
    \item Entropy: which uses the entropy of class probabilities to predict the uncertainty of samples.
    \revise{\item Confidence: which uses the highest probability score to predict the uncertainty of samples.}
    \item VAAL \cite{vaal}: which utilizes the variational autoencoder (VAE) to learn the feature space, and the discriminator to perform reverse training on the input data to determine whether it is labeled.
    \item DBAL \cite{DBAL}: which performs Bayesian CNNs to estimate the uncertainty of unlabeled points.
    \item coreset \cite{coreset}: which is a representative of distribution-based approaches and selects data that cover all highly diverse data according to the feature distribution.
    \item PAL \cite{PAL}: which adds a self-supervision head to the original classification network, and trains the self-supervision task in parallel with the original classification task.
    \item PT4AL\cite{PT4AL}: which sorts the unlabeled data in descending order of self-supervised task losses and divided into batches for each active learning cycle.  
    \revise{\item ActiveFT \cite{activeft}: which selects a subset of data with a similar distribution of the entire unlabeled pool and enough diversity by optimizing a parametric model.}
\end{enumerate} 


\subsubsection{Configuration.} Hot-start techniques (PT4AL\cite{PT4AL} and ours) label the initial samples through their respective techniques. Remaining techniques share an initial labeling sample set. We perform VGG16\cite{vgg} as backbones. For medium labeling budgets, to align the configuration, we follow the previous research \cite{PT4AL} and train the model from the weight in the latest cycle. For large labeling budgets, we train the model from scratch in each cycle to demonstrate the value of our sampled data. We calculate the average accuracy of the three runs. More details are \cref{tab:cfgs}. 

\begin{table*}[t]
\centering\small
\setlength{\tabcolsep}{4mm}
\begin{tabular}{c|c|cccccccc}
\toprule
\multirow{2}{*}{Dataset} & \multirow{2}{*}{Method} & \multicolumn{7}{c}{$\lambda$} \\
 &  & 10\% & 15\% & 20\% & 25\% & 30\% & 35\% & 40\% \\
 \midrule
\multirow{7}{*}{Caltech-101} & pt4al\cite{PT4AL} & 42.6 & 50.7 & 58.3 & 64.1 & 70.9 & 76.3 & 76.4 \\
 & pal\cite{PAL} & 45.2 & 53.4 & 56 & 61.6 & 65.8 & 69.4 & 75.9 \\
 & coreset\cite{coreset} & 45.2 & 53.2 & 56 & 61.3 & 63.2 & 65.7 & 71.9 \\
 & random & 45.3 & 49.6 & 56.8 & 60.4 & 63.6 & 64.6 & 68.4 \\
 & dbal\cite{DBAL} & 45.2 & 51.4 & 57.7 & 63.5 & 63.1 & 67.7 & 70.1 \\
 & \revise{activeft\cite{activeft}} & \revise{43.3} & \revise{51.3} & \revise{56.1} & \revise{60.1} & \revise{63.3} & \revise{65.4} & \revise{68.1} \\
 & \cellcolor{gray!20}bal & \cellcolor{gray!20}42.6\tiny{$\pm$0.0} & \cellcolor{gray!20}50.7\tiny{$\pm$0.3} & \cellcolor{gray!20}58.3\tiny{$\pm$0.2} & \cellcolor{gray!20}64.1\tiny{$\pm$0.7} & \cellcolor{gray!20}70.9\tiny{$\pm$0.3} & \cellcolor{gray!20}76.3\tiny{$\pm$0.7} & \cellcolor{gray!20}76.4\tiny{$\pm$0.6} \\
 \midrule
\multirow{8}{*}{SVHN} & pt4al\cite{PT4AL} & 87.9 & 91.5 & 92.4 & 92.5 & 93.7 & 94.6 & 94.9 \\
 & random & 86.1 & 89.1 & 90.1 & 90.5 & 91.6 & 91.9 & 92.2 \\
 & vaal\cite{vaal} & 86.1 & 88.1 & 88.9 & 90.1 & 91.2 & 91 & 91.7 \\
 & dbal\cite{DBAL} & 86.1 & 88.6 & 89.9 & 90.6 & 90.9 & 91.8 & 92.1 \\
 & pal\cite{PAL} & 86.1 & 89.8 & 91.3 & 92.4 & 93.4 & 93.6 & 93.9 \\
 & coreset\cite{coreset} & 86.1 & 89.6 & 91.3 & 92.4 & 93.1 & 93.5 & 93.7 \\
 & \revise{activeft\cite{activeft}} & \revise{87.8} & \revise{91.4} & \revise{92.3} & \revise{92.9} & \revise{93.3} & \revise{93.7} & \revise{93.8} \\
 & \cellcolor{gray!20}bal & \cellcolor{gray!20}89.1\tiny{$\pm$0.0} & \cellcolor{gray!20}92.6\tiny{$\pm$0.2} & \cellcolor{gray!20}93.3\tiny{$\pm$0.1} & \cellcolor{gray!20}93.6\tiny{$\pm$0.1} & \cellcolor{gray!20}94.4\tiny{$\pm$0.0} & \cellcolor{gray!20}94.8\tiny{$\pm$0.1} & \cellcolor{gray!20}95.0\tiny{$\pm$0.1} \\
 \midrule
\multirow{8}{*}{CIFAR-10} & pt4al\cite{PT4AL} & 71.9 & 81.9 & 84.5 & 86.9 & 88.6 & 90 & 90.9 \\
 & random & 64.3 & 67.4 & 71.8 & 74.5 & 77.4 & 78.5 & 79.8 \\
 & pal\cite{PAL} & 64.3 & 73.3 & 77.2 & 80.3 & 84 & 85.6 & 86.8 \\
 & coreset\cite{coreset} & 64.2 & 72.5 & 77.4 & 79.2 & 80.8 & 82.1 & 83.6 \\
 & dbal\cite{DBAL} & 64.3 & 73.1 & 77.6 & 81.5 & 82.4 & 84.7 & 85.8 \\
 & vaal\cite{vaal} & 64.3 & 68.4 & 73 & 76.8 & 78.7 & 79.8 & 80.8 \\
 & \revise{activeft\cite{activeft}} & \revise{43.0} & \revise{64.6} & \revise{70.3} & \revise{74.5} & \revise{76.2} & \revise{77.4} & \revise{77.8} \\
 & \cellcolor{gray!20}bal & \cellcolor{gray!20}68.7\tiny{$\pm$0.0} & \cellcolor{gray!20}83.6\tiny{$\pm$0.1} & \cellcolor{gray!20}86.5\tiny{$\pm$0.2} & \cellcolor{gray!20}88.2\tiny{$\pm$0.1} & \cellcolor{gray!20}89.8\tiny{$\pm$0.1} & \cellcolor{gray!20}90.7\tiny{$\pm$0.1} & \cellcolor{gray!20}91.2\tiny{$\pm$0.3} \\
 \midrule
\multirow{4}{*}{\revise{Tiny-ImageNet}} & \revise{pt4al\cite{PT4AL}} & \revise{7.9} & \revise{13.0} & \revise{17.8} & \revise{21.2} & \revise{23.5} & \revise{25.3} & \revise{26.3} \\
 & \revise{entropy} & \revise{9.9} & \revise{14.0} & \revise{17.5} & \revise{20.6} & \revise{23.3} & \revise{25.3} & \revise{26.8} \\
 & \revise{confidence} & \revise{9.9} & \revise{14.7} & \revise{18.0} & \revise{21.3} & \revise{23.7} & \revise{25.6} & \revise{27.3} \\
 & \cellcolor{gray!20}\revise{bal} & \cellcolor{gray!20}\revise{10.3\tiny{$\pm$0.0}} & \cellcolor{gray!20}\revise{15.5\tiny{$\pm$0.2}} & \cellcolor{gray!20}\revise{19.6\tiny{$\pm$0.2}} & \cellcolor{gray!20}\revise{22.6\tiny{$\pm$0.2}} & \cellcolor{gray!20}\revise{24.6\tiny{$\pm$0.1}} & \cellcolor{gray!20}\revise{26.5\tiny{$\pm$0.0}} & \cellcolor{gray!20}\revise{28.0\tiny{$\pm$0.1}}\\
\bottomrule
\end{tabular}
\caption{\textbf{Medium labeling budget}: Experiments on CIFAR-10, SVHN, Caltech-101, \revise{and Tiny-ImageNet}, comparing random sampling, entropy sampling, VAAL \cite{vaal}, DBAL \cite{DBAL}, coreset \cite{coreset}, PAL \cite{PAL}, PT4AL \cite{PT4AL}, \revise{ActiveFT \cite{activeft}}, and our proposed BAL.}
\label{tab:med}
\end{table*}

\begin{figure*}[t]
\centering
\subfloat[CIFAR-10]{\includegraphics[width=0.25\linewidth,trim=0 0 0 0,clip]{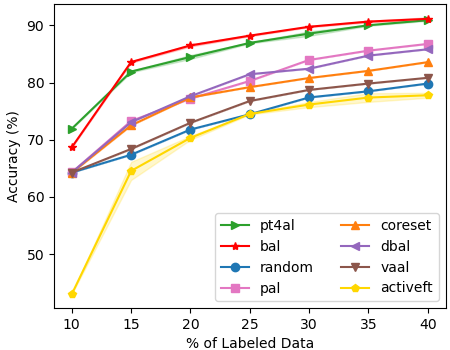}}
\subfloat[SVHN]{\includegraphics[width=0.25\linewidth,trim=0 0 0 0,clip]{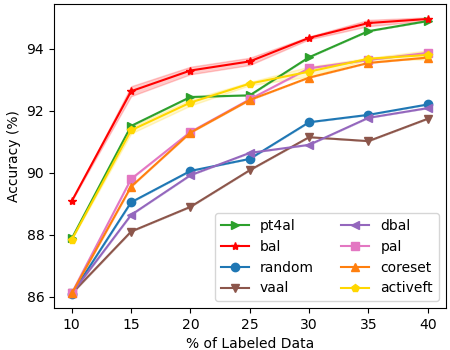}}
\subfloat[Caltech-101]{\includegraphics[width=0.25\linewidth,trim=0 0 0 0,clip]{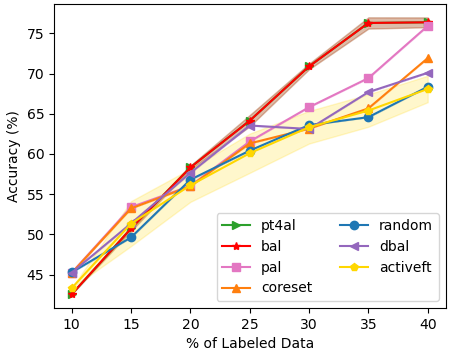}}
\subfloat[\revise{Tiny-ImageNet}]{\includegraphics[width=0.25\linewidth,trim=0 0 0 0,clip]{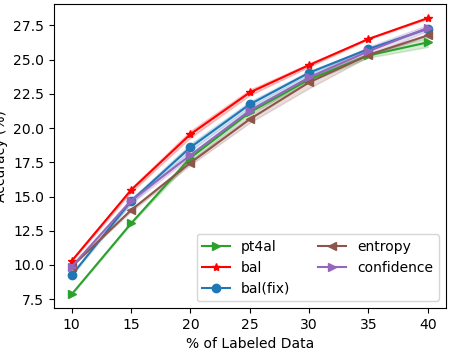}}
\caption{\textbf{Medium labeling budgets}: Experiments on (a) CIFAR-10, (b) SVHN, (c) Caltech-101, \revise{and (d) Tiny-ImageNet}. Comparing methods include random sampling, entropy sampling, VAAL \cite{vaal}, DBAL \cite{DBAL}, Coreset \cite{coreset}, PAL \cite{PAL}, PT4AL \cite{PT4AL}, \revise{ActiveFT \cite{activeft}}, and proposed BAL. \revise{We do not perform ActiveFT on Tiny-ImageNet due to its excessive memory requirements.} Light colors represent the variance. }
\label{fig:med}
\end{figure*}

\begin{table*}[t]
\centering
\small
\setlength{\tabcolsep}{1.7mm}
\begin{tabular}{c|c|ccccccccccccc}
\toprule
\multirow{2}{*}{Dataset} & \multirow{2}{*}{Method} & \multicolumn{10}{c}{$\lambda$} \\
 &  & 10\% & 20\% & 30\% & 40\% & 50\% & 60\% & 70\% & 80\% & 90\% & 100\% \\
  \midrule
\multirow{3}{*}{Caltech-101} & pt4al\cite{PT4AL} & 50.7 & 61.4 & 68.1 & 71.1 & 73.6 & 76.8 & 80.5 & 82.6 & 83.2 & 83.6 \\
 & \revise{activeft\cite{activeft}} & \revise{51.5} & \revise{59.7} & \revise{65.6} & \revise{70.1} & \revise{73.9} & \revise{76.8} & \revise{79.0} & \revise{81.3} & \revise{82.6} & \revise{83.9} \\
 & \cellcolor{gray!20}bal & \cellcolor{gray!20}40.1\tiny{$\pm$0.8} & \cellcolor{gray!20}56.1\tiny{$\pm$0.8} & \cellcolor{gray!20}64.7\tiny{$\pm$0.5} & \cellcolor{gray!20}71.4\tiny{$\pm$0.2} & \cellcolor{gray!20}78.1\tiny{$\pm$1.0} & \cellcolor{gray!20}85.4\tiny{$\pm$0.4} & \cellcolor{gray!20}90.9\tiny{$\pm$0.6} & \cellcolor{gray!20}92.8\tiny{$\pm$0.5} & \cellcolor{gray!20}93.0\tiny{$\pm$0.5} & \cellcolor{gray!20}92.7\tiny{$\pm$0.4} \\
  \midrule
\multirow{3}{*}{SVHN} & pt4al\cite{PT4AL} & 88.5 & 90.9 & 92.1 & 93 & 93.5 & 93.8 & 94.1 & 95.1 & 95.3 & 95.6 \\
 & \revise{activeft\cite{activeft}} & \revise{88.4} & \revise{92.1} & \revise{93.1} & \revise{93.9} & \revise{94.3} & \revise{94.6} & \revise{94.7} & \revise{94.8} & \revise{94.9} & \revise{95.1} \\
 & \cellcolor{gray!20}bal & \cellcolor{gray!20}89.1\tiny{$\pm$0.0} & \cellcolor{gray!20}92.1\tiny{$\pm$0.0} & \cellcolor{gray!20}93.7\tiny{$\pm$0.0} & \cellcolor{gray!20}94.2\tiny{$\pm$0.0} & \cellcolor{gray!20}94.8\tiny{$\pm$0.0} & \cellcolor{gray!20}95.3\tiny{$\pm$0.0} & \cellcolor{gray!20}95.7\tiny{$\pm$0.0} & \cellcolor{gray!20}95.8\tiny{$\pm$0.0} & \cellcolor{gray!20}95.7\tiny{$\pm$0.0} & \cellcolor{gray!20}95.6\tiny{$\pm$0.1} \\
  \midrule
\multirow{3}{*}{CIFAR-10} & pt4al\cite{PT4AL} & 53.6 & 80.1 & 84.5 & 85.9 & 87.3 & 88.9 & 90.4 & 91.4 & 91.8 & 92.2 \\
 & \revise{activeft\cite{activeft}} & \revise{45.1} & \revise{70.1} & \revise{75.5} & \revise{77.4} & \revise{79.6} & \revise{79.4} & \revise{81.4} & \revise{81.6} & \revise{82.1} & \revise{83.0} \\
 & \cellcolor{gray!20}bal & \cellcolor{gray!20}54.4\tiny{$\pm$2.6} & \cellcolor{gray!20}72.7\tiny{$\pm$0.2} & \cellcolor{gray!20}77.4\tiny{$\pm$0.1} & \cellcolor{gray!20}80.2\tiny{$\pm$0.2} & \cellcolor{gray!20}82.2\tiny{$\pm$0.3} & \cellcolor{gray!20}83.4\tiny{$\pm$0.3} & \cellcolor{gray!20}85.0\tiny{$\pm$0.1} & \cellcolor{gray!20}85.0\tiny{$\pm$0.3} & \cellcolor{gray!20}85.3\tiny{$\pm$0.1} & \cellcolor{gray!20}85.2\tiny{$\pm$0.1} \\
 \midrule
\multirow{2}{*}{\revise{Tiny-ImageNet}} & \revise{pt4al\cite{PT4AL}} & \revise{7.3} & \revise{13.6} & \revise{20.0} & \revise{24.1} & \revise{27.5} & \revise{30.1} & \revise{32.2} & \revise{34.3} & \revise{35.2} & \revise{36.5} \\
 & \cellcolor{gray!20}\revise{bal} & \cellcolor{gray!20}\revise{9.4\tiny{$\pm$0.1}} & \cellcolor{gray!20}\revise{17.1\tiny{$\pm$0.2}} & \cellcolor{gray!20}\revise{23.8\tiny{$\pm$0.2}} & \cellcolor{gray!20}\revise{28.2\tiny{$\pm$0.2}} & \cellcolor{gray!20}\revise{31.1\tiny{$\pm$0.3}} & \cellcolor{gray!20}\revise{32.7\tiny{$\pm$0.2}} & \cellcolor{gray!20}\revise{34.4\tiny{$\pm$0.0}} & \cellcolor{gray!20}\revise{35.5\tiny{$\pm$0.2}} & \cellcolor{gray!20}\revise{37.1\tiny{$\pm$0.2}} & \cellcolor{gray!20}\revise{37.2\tiny{$\pm$0.1}}\\
\bottomrule
\end{tabular}
\caption{\textbf{Large labeling budget}: Experiments on CIFAR-10, SVHN, Caltech-101 and \revise{Tiny-ImageNet}, comparing BAL with the current SOTAs, PT4AL \cite{PT4AL} \revise{and ActiveFT \cite{activeft}}. In order to compare with the data trained on the full dataset fairly, we train from scratch in each cycle.}
\label{tab:large}
\end{table*}

\begin{figure*}[t]
\centering
\subfloat[CIFAR-10]{\includegraphics[width=0.25\linewidth]{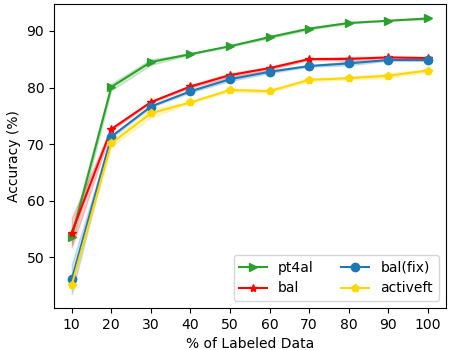}}
\subfloat[SVHN]{\includegraphics[width=0.25\linewidth]{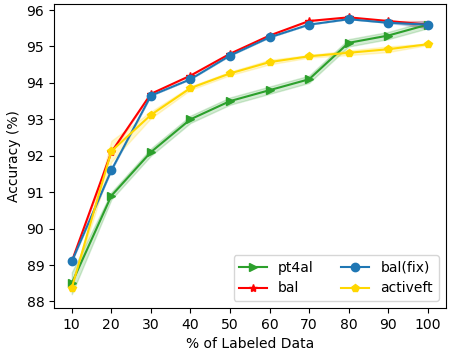}}
\subfloat[Caltech-101]{\includegraphics[width=0.25\linewidth]{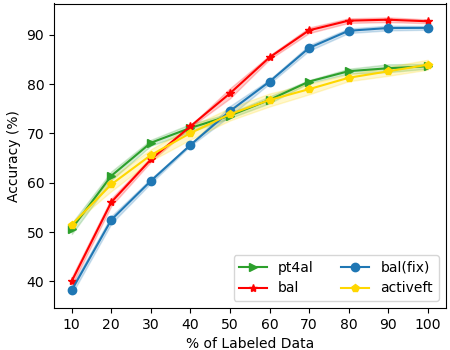}}
\subfloat[\revise{Tiny-ImageNet}]{\includegraphics[width=0.25\linewidth]{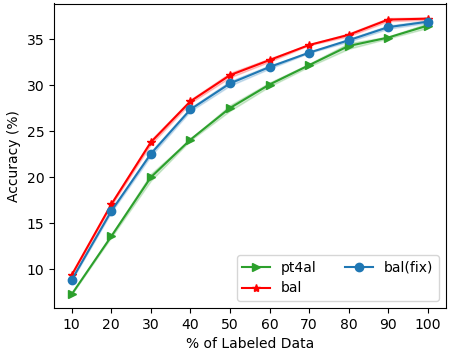}}
\caption{\textbf{Large labeling budgets}: Experiments on (a) CIFAR-10, (b) SVHN, (c) Caltech-101 \revise{and (d) Tiny-ImageNet}. Comparing methods include BAL and the current SOTAs, PT4AL \cite{PT4AL} \revise{and ActiveFT \cite{activeft}}. We train from scratch in each cycle. Light colors represent the variance. Small figures show the accuracy when labeling 80\% of the datasets.}
\label{fig:large}
\end{figure*}

\subsubsection{Proportion of labeled data.} The evaluation criteria for active learning methods are usually the accuracy when a fixed proportion of labeled data $\lambda=\frac{IK}{N}$ is used, where $N$ is the number of the dataset; $I$ is the number of active learning cycles; $K$ is the number of labeled samples in each cycle. We verify our approaches threefold:
\begin{enumerate}
    \item First, we verify our approach in a commonly-acknowledge benchmark - The initial pool accounts for 10\% of the entire dataset, and an additional 5\% of the dataset selected through various active learning techniques is added to each query until it reaches 40\%.
    \item Then, we test our method for large labeling budgets. The initial pool labels 10\%, and we add 10\% in each cycle until it reaches 100\%, where the model on 100\% is a normal classification task. 
    \revise{\item Finally, we perform our method for small labeling budgets. The initial pool comprises 2\% of the total dataset, and with each query, an extra 2\% of the dataset is incrementally incorporated using diverse active learning methods until it reaches a total of 10\%.}
\end{enumerate}
\revise{In summary, our approach has been extensively validated across a range of datasets and labeling budgets, and compared with various approaches, ensuring its robustness and versatility.}

\begin{table*}[t]
\centering
\small
\setlength{\tabcolsep}{6.5mm}
\begin{tabular}{c|c|cccccc}
\toprule
\multirow{2}{*}{Dataset} & \multirow{2}{*}{Method} & \multicolumn{5}{c}{$\lambda$} \\
 & & 2\% & 4\% & 6\% & 8\% & 10\% \\
  \midrule
\multirow{5}{*}{Caltech-101} & pt4al\cite{PT4AL} & 21.6 & 31.6 & 34.8 & 38.2 & 42.3 \\
 & entropy & 31.4 & 32.9 & 36.4 & 40.5 & 43.4 \\
 & confidence & 34.9 & 34.8 & 39.4 & 41.9 & 45 \\
 & \revise{activeft\cite{activeft}} & \revise{34.1} & \revise{36.3} & \revise{39.9} & \revise{44.4} & \revise{48.8} \\
 & \cellcolor{gray!20}bal & \cellcolor{gray!20}33.1\tiny{$\pm$0.6} & \cellcolor{gray!20}38.7\tiny{$\pm$2.3} & \cellcolor{gray!20}43.1\tiny{$\pm$0.3} & \cellcolor{gray!20}47.4\tiny{$\pm$1.2} & \cellcolor{gray!20}50.8\tiny{$\pm$1.2} \\
  \midrule
\multirow{5}{*}{SVHN} & pt4al\cite{PT4AL} & 15.7 & 15.9 & 50.4 & 71.7 & 87.2 \\
 & entropy & 32.5 & 39.3 & 62.3 & 88.9 & 91.1 \\
 & confidence & 19.6 & 29.7 & 82.2 & 89.7 & 91.7 \\
 & \revise{activeft\cite{activeft}} & \revise{25.6} & \revise{55.3} & \revise{85.4} & \revise{88.0} & \revise{89.2} \\
 & \cellcolor{gray!20}bal & \cellcolor{gray!20}31.9\tiny{$\pm$0.0} & \cellcolor{gray!20}81.3\tiny{$\pm$5.4} & \cellcolor{gray!20}88.6\tiny{$\pm$1.4} & \cellcolor{gray!20}90.9\tiny{$\pm$0.6} & \cellcolor{gray!20}92.1\tiny{$\pm$0.3} \\
  \midrule
\multirow{5}{*}{CIFAR-10} & pt4al\cite{PT4AL} & 12.2 & 12.9 & 24.9 & 41.4 & 55.1 \\
 & entropy & 14.6 & 25.7 & 53.4 & 64 & 68.6 \\
 & confidence & 11.9 & 16.7 & 40.6 & 58.1 & 64.4 \\
 & \revise{activeft\cite{activeft}} & \revise{12.7} & \revise{20.8} & \revise{42.0} & \revise{53.1} & \revise{59.0} \\
 & \cellcolor{gray!20}bal & \cellcolor{gray!20}15.3\tiny{$\pm$1.3} & \cellcolor{gray!20}36.7\tiny{$\pm$3.1} & \cellcolor{gray!20}55.5\tiny{$\pm$2.9} & \cellcolor{gray!20}63.2\tiny{$\pm$1.2} & \cellcolor{gray!20}67.9\tiny{$\pm$0.4} \\
 \midrule
\multirow{4}{*}{\revise{Tiny-ImageNet}} & \revise{pt4al\cite{PT4AL}} & \revise{2.6} & \revise{4.1} & \revise{5.9} & \revise{7.4} & \revise{8.6} \\
 & \revise{entropy} & \revise{2.2} & \revise{3.4} & \revise{4.7} & \revise{5.6} & \revise{6.8} \\
 & \revise{confidence} & \revise{2.2} & \revise{3.7} & \revise{5.1} & \revise{6.5} & \revise{7.7} \\
 & \cellcolor{gray!20}\revise{bal} & \cellcolor{gray!20}\revise{2.6\tiny{$\pm$0.1}} & \cellcolor{gray!20}\revise{5.0\tiny{$\pm$0.1}} & \cellcolor{gray!20}\revise{6.3\tiny{$\pm$0.2}} & \cellcolor{gray!20}\revise{7.9\tiny{$\pm$0.2}} & \cellcolor{gray!20}\revise{9.1\tiny{$\pm$0.3}}\\
\bottomrule
\end{tabular}
\caption{\textbf{Small labeling budget}: Experiments on CIFAR-10, SVHN, Caltech-101 and Tiny-ImageNet, comparing BAL with the current SOTAs, PT4AL \cite{PT4AL} and ActiveFT \cite{activeft}.}
\label{tab:small}
\end{table*}

\begin{figure*}[t]
\centering
\subfloat[CIFAR-10]{\includegraphics[width=0.25\linewidth,trim=0 0 0 0,clip]{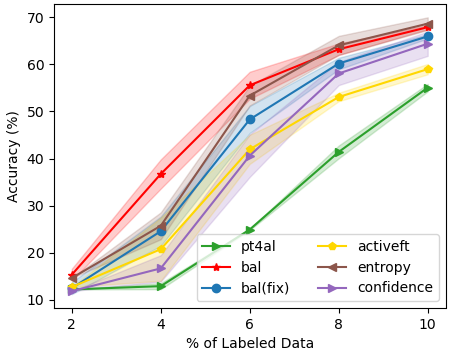}}
\subfloat[SVHN]{\includegraphics[width=0.25\linewidth,trim=0 0 0 0,clip]{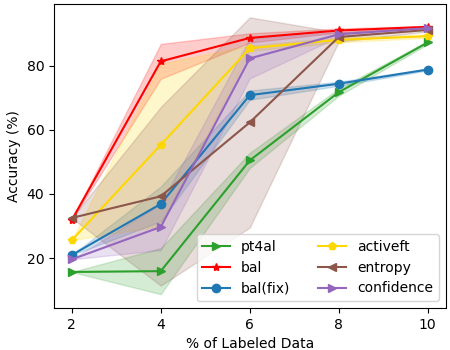}}
\subfloat[Caltech-101]{\includegraphics[width=0.25\linewidth,trim=0 0 0 0,clip]{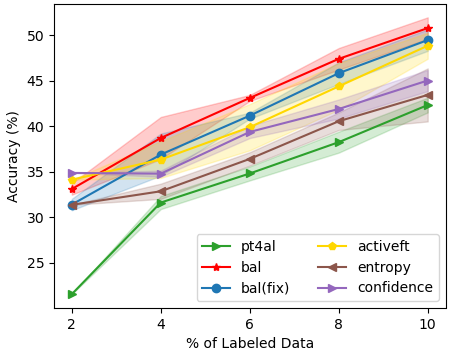}}
\subfloat[\revise{Tiny-ImageNet}]{\includegraphics[width=0.25\linewidth,trim=0 0 0 0,clip]{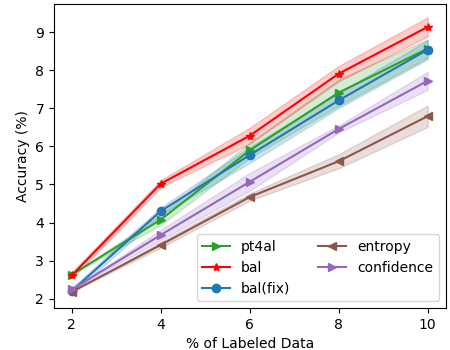}}
\caption{\textbf{Small labeling budgets}: Experiments on (a) CIFAR-10, (b) SVHN, (c) Caltech-101 \revise{and (d) Tiny-ImageNet}. Comparing methods include random sampling, entropy sampling, confidence sampling, PT4AL \cite{PT4AL}, \revise{ActiveFT \cite{activeft}}, and proposed BAL. Light colors represent the variance. }
\label{fig:small}
\end{figure*}

\begin{figure*}[t]
\centering
\subfloat[Random sampling]{\includegraphics[width=\linewidth,trim=640 570 290 80,clip]{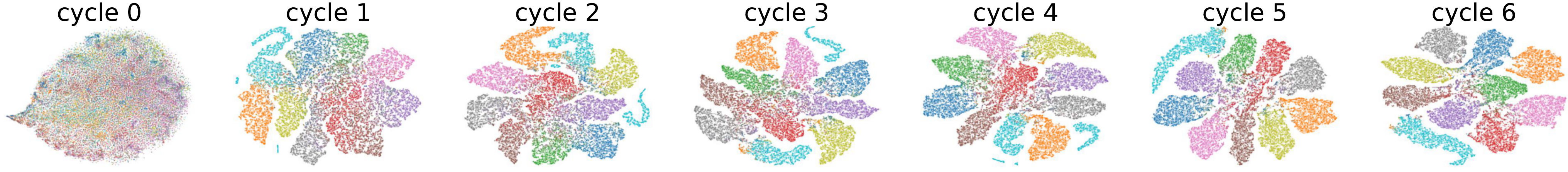}}
\hfil
\subfloat[PT4AL]{\includegraphics[width=\linewidth,trim=640 340 290 310,clip]{figures/exp_vis.pdf}}
\hfil
\subfloat[BAL (ours)]{\includegraphics[width=\linewidth,trim=640 110 290 540,clip]{figures/exp_vis.pdf}}
\caption{The t-SNE visualization of features learned by (a) Random sampling, (b) PT4AL \cite{PT4AL}, and (c) proposed BAL on CIFAR-10 in different cycles. Bright colors represent labeled samples.}
\label{fig:exp_vis}
\end{figure*}

\subsection{Medium labeling budgets}
\label{sec:exp_med}
Firstly, we compare the performance of uniformly-divided sub-pools \cite{PT4AL} and adaptive sub-pools, where balancing factor $\beta=1.3$ is chosen according to the aforementioned mechanism (\cref{alg1}). Results are illustrated in \cref{fig:beta}. Our adaptive sub-pools outperform the other consistently on three datasets. We note that when performing on Caltech-10l, where the categories distribute unevenly, adaptive sub-pools show great prior - Its accuracy is 1.0\% higher than naive sub-pools. Next, we test our approach for standard medium labeling budgets. Results are shown in \cref{tab:small} and illustrate in \cref{fig:med}. We compares the performance of various techniques with different labeled proportion $\lambda$. It shows that BAL outperforms other methods with large margins. On CIFAR-10, performance of BAL with $\lambda=15\%$ is close to that of the current SOTA, PT4AL \cite{PT4AL}, with $\lambda=20\%$, saving 25\% of the data labeling. On SVHN, BAL with $\lambda=15\%$ even defeats  PT4AL with $\lambda=25\%$. On Caltech-101, BAL boosts the result of the current SOTA by 5.4\% at $\lambda=35\%$. 


\subsection{Large Labeling Budgets}
\label{sec:exp_large}
In this section, we validate our method for large budgets. The sampler $\psi$ needs to select more data in each sub-pool due to a larger labeled proportion $\lambda$. In this way, if the same balancing factor $\beta$ is performed, the sampled data will be generally determined by sub-pools $\{P_U^i\}_{i=1}^I$, while the impact of uncertainty becomes less. Thus, the sampled data will be more diverse but less uncertain. In order to balance diversity and uncertainty again, BAL set a higher balancing factor $\beta=2.0$ to obtain a large $|P_U^i|$ in each cycle and thus ensure sufficient choices of the uncertainty-based sampler. Results are shown in \cref{tab:large} and illustrate in \cref{fig:large}. It shows that the performance based on 80\% samples selected by the current SOTA, PT4AL \cite{PT4AL} is 0.74\% lower than that on the full dataset, while our approach is on par with the latter. It shows that BAL can sample valuable data which is able to offer comparative information with the whole dataset. Thus, we only need to train on our selected partial dataset and save 20\% labeling budget without performance deterioration.

\begin{figure*}[t]
\centering
\subfloat[Randomly dividing sub-pool]{\includegraphics[width=0.33\linewidth]{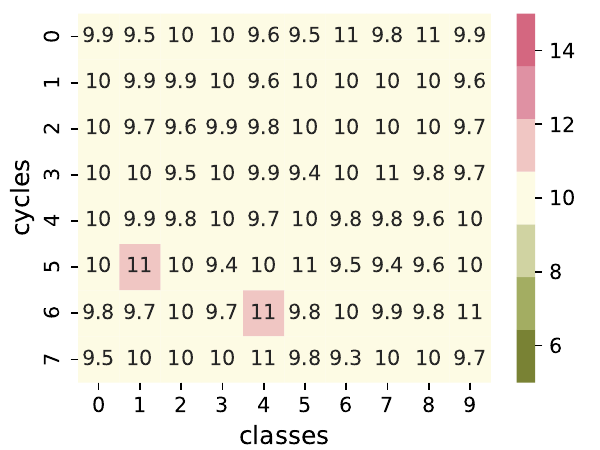}}
\subfloat[uniformly-divided \cite{PT4AL}]{\includegraphics[width=0.33\linewidth]{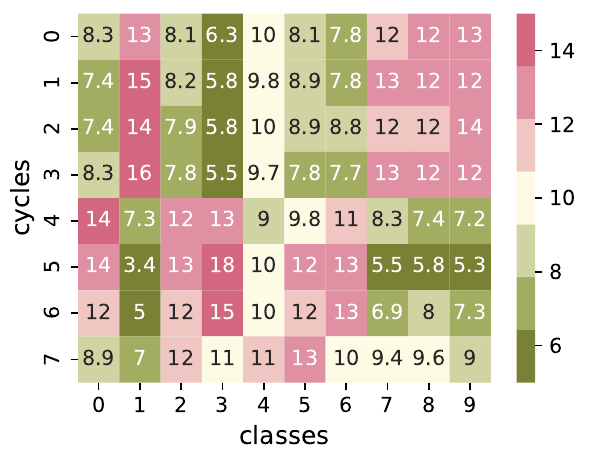}}
\subfloat[Adaptive (ours)]{\includegraphics[width=0.33\linewidth]{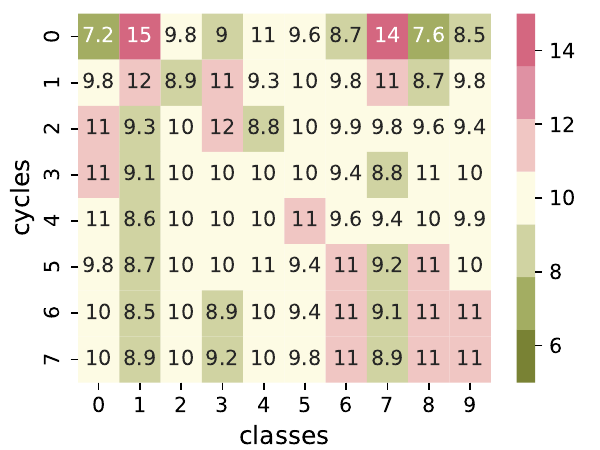}}
\caption{The number of classes in (a) uniformly-divided \cite{PT4AL} and (b) adaptive sub-pools on CIFAR-10.}
\label{fig:cls_distribution}
\end{figure*}

\subsection{\revise{Small labeling budgets}}
\label{sec:exp_small}
\revise{The prior theoretical analysis \cite{typiclust}, indicates a behavior analogous to a phase transition phenomenon: under limited budget conditions, it is more beneficial to query typical examples, while in situations with a larger budget, the most effective strategy is to query atypical or unrepresentative examples. Therefore, to further highlight the resilience of our Balancing Active Learning approach, we conduct experiments with limited labeling budgets. The outcomes, displayed in \cref{tab:small} and visually depicted in \cref{fig:small}, involve a performance comparison among diverse techniques across varying labeled proportions $\lambda$. Notably, BAL consistently performs outstandingly even when the current SOTAs in common benchmarks, PT4AL~\cite{PT4AL} and ActiveFT~\cite{activeft}, fail. When training on $4\%$ of the total dataset, BAL outperforms the second-best method by 11\% on CIFAR-10 and by 26\% on SVHN.
}
\subsection{Visualization}
\label{sec:vis}
We illustrate the t-SNE visualization of features for vivid exhibition in \cref{fig:exp_vis}. The samples are labeled by random sampling, PT4AL\cite{PT4AL} and our proposed BAL on CIFAR-10. In cycle 0, features of different categories are almost mixed evenly when randomly sampled. PT4AL improves results of the baseline, while our approach distinguishes different classes with clearer boundaries. In the last cycle, although both PT4AL and ours can separate different categories, features from the same class are gathered more tightly when performing BAL. In \cref{fig:cls_distribution}, we illustrate the category distribution of the uniformly-divided sub-pools in PT4AL \cite{PT4AL} and adaptive sub-pools in BAL in each active learning cycle. It has shown that our sub-pools divide classes more evenly compared with PT4AL. Experiments of \cite{PT4AL} have shown that an imbalance in the distribution adversely affects the results. Therefore, the balance of uncertainty and diversity assists in extracting more even data in each cycle, which favors active learning tasks.

%% file: sections/06_ablation.tex
In this section, we demonstrate our exploration of several design options for BAL. The experiments are performed on CIFAR-10 with the same configuration in \cref{sec:exp}. In subsection \cref{sec:abss}, we discuss two ways to set labels for the self-supervising task; in subsection \cref{sec:abcdd}, we discuss two criteria for dividing batches, including SoC and DtC; in subsection \cref{sec:absort}, we discuss the order of SoC sorting, including from largest to smallest or  from smallest to largest; and in subsection \cref{sec:absampling}, we discuss sampling methods, including confidence, entropy, Kmeans clustering, and random sampling. Of all the methods, we chose the one that works best as the component of the SSAL framework, i.e. the red curve in \cref{fig:ablation}. 

\subsection{Self-supervised Task.} 
\label{sec:abss}
\revise{The results presented in \cref{fig:ab_ss} shed light on the effectiveness of using a straightforward rotation prediction pretext task for self-supervised pretext task for active learning. In contrast to the latest and more complex self-supervised learning approaches like SimCLR \cite{simclr}, the relatively simpler task of predicting image rotations yields superior outcomes when incorporated into the active learning framework. This finding underscores the notion that, in certain scenarios, a less intricate pretext task can outperform more advanced alternatives, highlighting the significance of task selection in the design of active learning strategies.}

We define two rotation losses, the first loss function of which is defined in \cref{equ:L_SS}, where $y_{\theta}$ represents labels corresponding to angle $\theta$. That is, when images rotate the same angle $\theta$, their corresponding labels are the same ($y_{\theta}$). Another way to define the pretext task is:
\begin{equation}
L(x_i)=\frac{1}{k}\sum_{\theta\in\{0,90,180,270\}}L_{CE}(l_{ss}(t(x_i, \theta)), y_{i})
\end{equation} 
where $L_{CE}$ is the cross-entropy loss. $t(x_i, \theta)$ represents an image that rotates the input image $x_i$ by $\theta$ degree. $l_{ss}(\cdot)$ represents output of self-supervised model $F_{ss}(\cdot)$. $y_i$ represents labels corresponding to the $i$-th image $x_i$. That is, the labels of an image $x_i$ (although rotates at different angles) are the same. As seen from \cref{fig:ab_ss}, labels according to rotation angles perform better. 

\subsection{Metric.} 
\label{sec:abcdd}
\begin{figure*}[!t]
\centering
\subfloat[\revise{Self-Supervised Task}]{\includegraphics[width=0.245\linewidth]{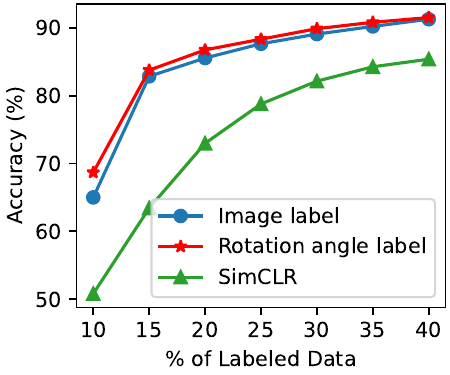}\label{fig:ab_ss}}
\hfil
\subfloat[Metric]{\includegraphics[width=0.245\linewidth]{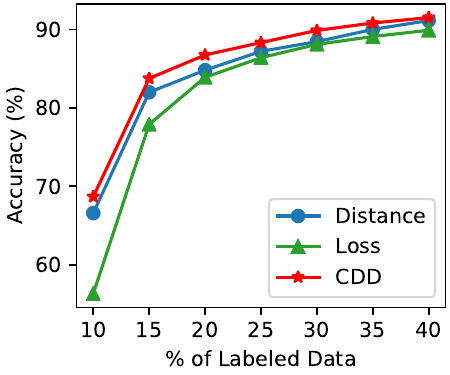}\label{fig:ab_metrics}}
\subfloat[Sorting]{\includegraphics[width=0.245\linewidth]{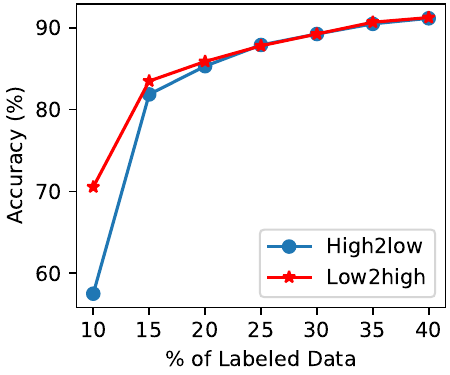}\label{fig:ab_sorted}}
\hfil
\subfloat[Sampling]{\includegraphics[width=0.245\linewidth]{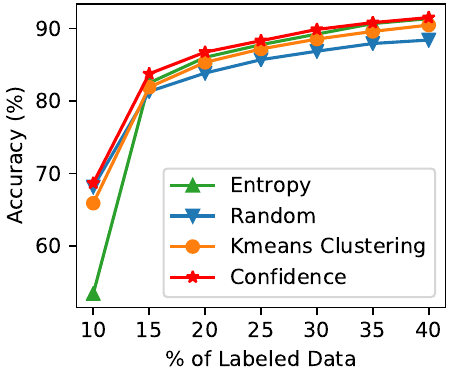}\label{fig:ab_sampling}}
\caption{Ablation Experiments on CIFAR-10. (a) Self-supervising tasks, including \revise{SimCLR \cite{simclr}}, and rotation prediction with labels according to images or rotation angles. (b) Metrics for dividing batches, including the CDD and the distance to the nearest cluster center. (c) Orders of batch sorting, including in ascending and descending order. (d) Methods of sampling, including confidence, entropy, K-means clustering, and random sampling.}
\label{fig:ablation}
\end{figure*}

In \cref{sec:ss}, we propose two criteria for sorting the sub-pools $\{P_U^i\}_{i=1}^I$, namely our proposed CDD and a naive baseline, the distance to the nearest cluster center. Results are shown in \cref{fig:ab_metrics}, which proves that CDD works better. We also compare CDD with another indicator, loss of pretext task, of the current SOTA, PT4AL \cite{PT4AL}. CDD outperforms it by 1.36\%.

\subsection{Sorting.} 
\label{sec:absort}
When sorting CDDs, we look at two different sorting strategies, that is, in ascending or descending order. For both strategies, in the first cycle, the $K$ samples are at the top of the sorted samples. As shown in \cref{fig:ab_sorted}, the performance in ascending order performs better. Thus it is better to learn difficult samples first. What's more, the results of the two strategies tend to be consistent when $\lambda$ is larger. It shows these sorting methods have a more significant impact on data initialization.

\subsection{Sampling.} 
\label{sec:absampling}
We explore four different sampling methods, including (i) a confidence-based sampler, which selects data with the lowest posterior probability, (ii) an entropy-based sampler, which selects data with the highest entropy, (iii) a clustering-based sampler, which K-means clusters features and samples the data closest to cluster centers, and (iv) a random sampler. As shown in \cref{fig:ab_sampling}, the confidence-based sampler performs the best.

%% file: sections/07_limitation.tex
\revise{In this paper, we concentrate on providing a comprehensive experimental validation in the realm of image-level classification tasks. In this scene, we have observed that simple rotation prediction methods can outperform more advanced pretext tasks such as SimCLR \cite{simclr}. On the other hand, for pixel-level tasks like segmentation, we consider pixel-level reconstruction tasks like MAE \cite{mae} and BEiT \cite{beit} more suitable. Nevertheless, these tasks typically are performed on Transformer \cite{vit} frameworks, which differ from the typical benchmarks and previous approaches, and may lead to unfair comparison. Due to these constraints, we do not further explore them in this paper. However, we are confident that the concepts of Cluster Distance Difference and balancing novelty and diversity proposed in this work hold potential for further extension to various downstream tasks and real-world applications. Based on the aforementioned ideas, we will continuously dedicate ourselves to further expanding its applicability to a broader spectrum of downstream tasks and real-world applications. We hope that these insights will also inspire future researchers to make further advancements in active learning.}

%% file: sections/08_conclusion.tex
In this paper, we utilize the K-means clustering on features obtained by self-supervised learning. Then we design a diversity-based indicator, the cluster distance difference (CDD), which leverages the information of the inter-samples relationship to benefit downstream tasks. Furthermore, we propose the Balancing Active Learning (BAL) framework to balance diverse data with uncertain data by proposed adaptive sub-pools. Our approach surpasses all previous active learning methods in commonly-acknowledged benchmarks by significant margins. At last, we experimentally verified that when labeling 80\% samples, the performance of the current SOTA deteriorates while our proposed BAL remains. It proves that BAL has the potential to save budgets as well as reach comparative performance.

%% file: sections/09_bio.tex
\begin{IEEEbiography}
 [{\includegraphics[height=1.10in,clip,keepaspectratio]{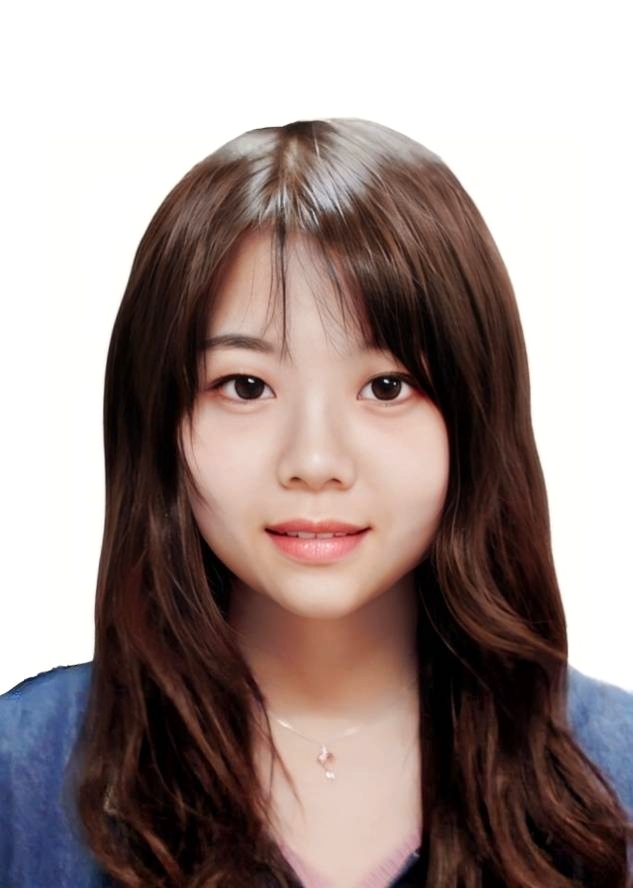}}]
 {Jingyao Li} received the B.Eng. degree from Xi'an Jiaotong University. She is currently a Ph.D. student at Department of Computer Science and Engineering of the Chinese University of Hong Kong (CUHK), under the supervision of Prof. Jiaya Jia. Her research interests include self-supervised learning, knowledge distillation and out-of-distribution detection.
\end{IEEEbiography}

\begin{IEEEbiography}
 [{\includegraphics[height=1.25in,clip,keepaspectratio]{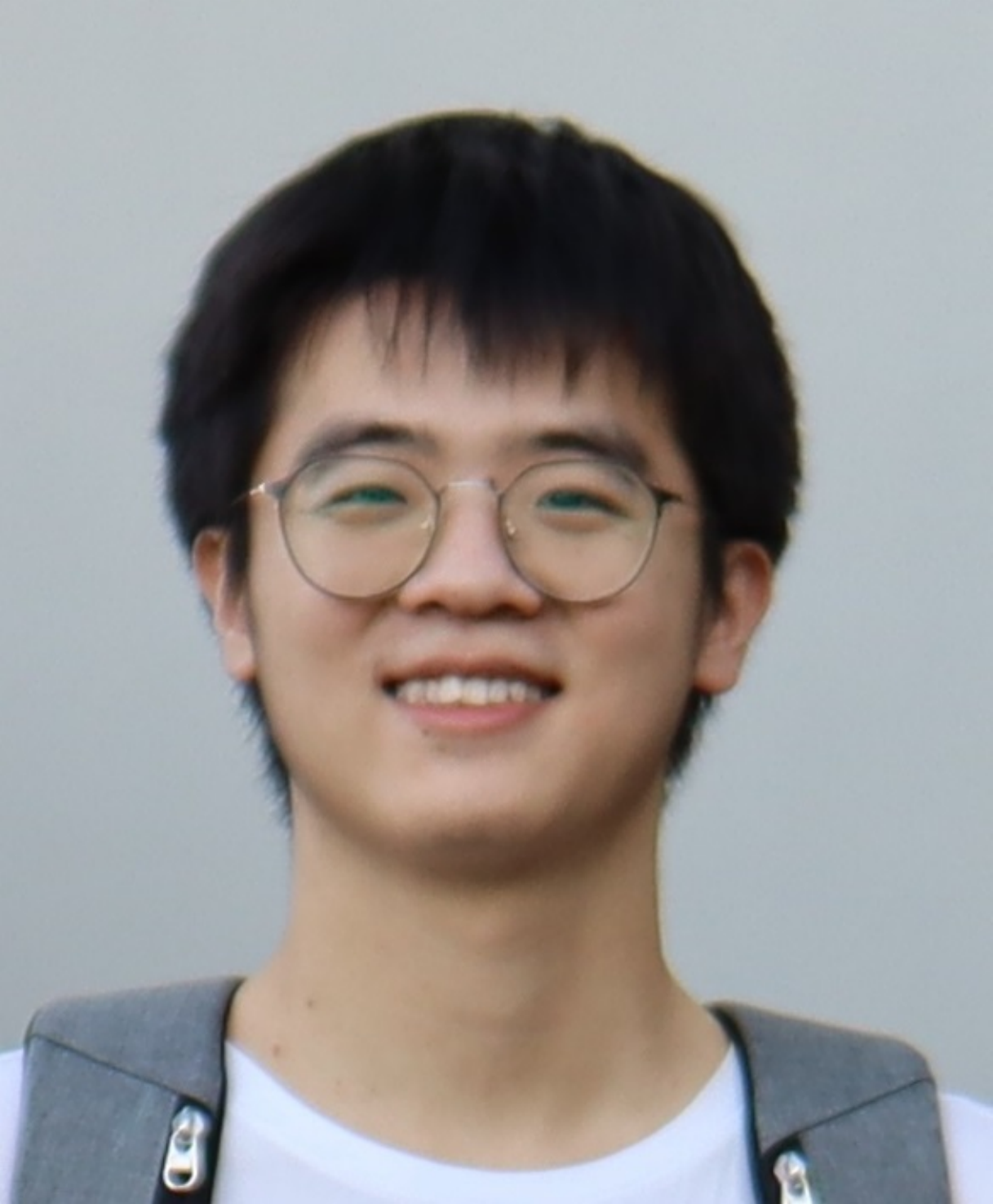}}] {Pengguang Chen} received the B.Eng. degree in Computer Science from Nanjing University and the Ph.D. degree from the Chinese University of Hong Kong (CUHK), under the supervision of Prof. Jiaya Jia. He is currently a researcher in SmartMore. He serves as a reviewer for CVPR, ICCV, ECCV, TPAMI. His research interests include neural architecture search, self-supervised learning, knowledge distillation and semantic segmentation.
\end{IEEEbiography}

\begin{IEEEbiography}[{\includegraphics[width=1in,height=1.10in,clip,keepaspectratio]{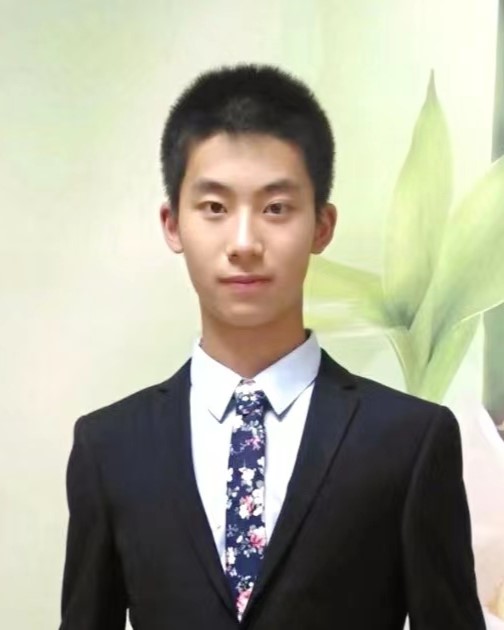}}]{Shaozuo Yu} is a Ph.D. student at Department of Computer Science and Engineering of the Chinese University of Hong Kong. He served as a program chair of the workshop and challenge on “Out-of-Distribution Generalization in Computer Vision” at ECCV’22. He served as a reviewer for CVPR, Neurips, and ICML. His research interests include multimodality, generative models, and robust vision.
\end{IEEEbiography}

\begin{IEEEbiography}
[{\includegraphics[width=1in,height=1.25in,clip,keepaspectratio]{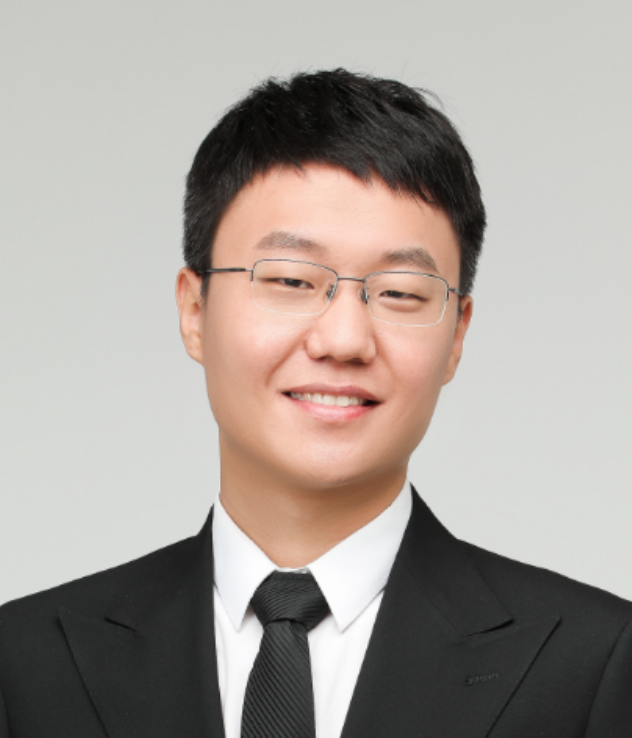}}]{Shu Liu} now serves as Co-Founder and Technical Head in SmartMore. He received the BS degree from Huazhong University of Science and Technology and the PhD degree from the Chinese University of Hong Kong. He was the winner of 2017 COCO Instance Segmentation Competition and received the Outstanding Reviewer of ICCV in 2019. He continuously served as a reviewer for TPAMI, CVPR, ICCV, NIPS, ICLR and etc. His research interests lie in deep learning and computer vision. 
  He is a member of IEEE.
 \end{IEEEbiography} 

\begin{IEEEbiography}
[{\includegraphics[width=1in,height=1.25in,clip,keepaspectratio]{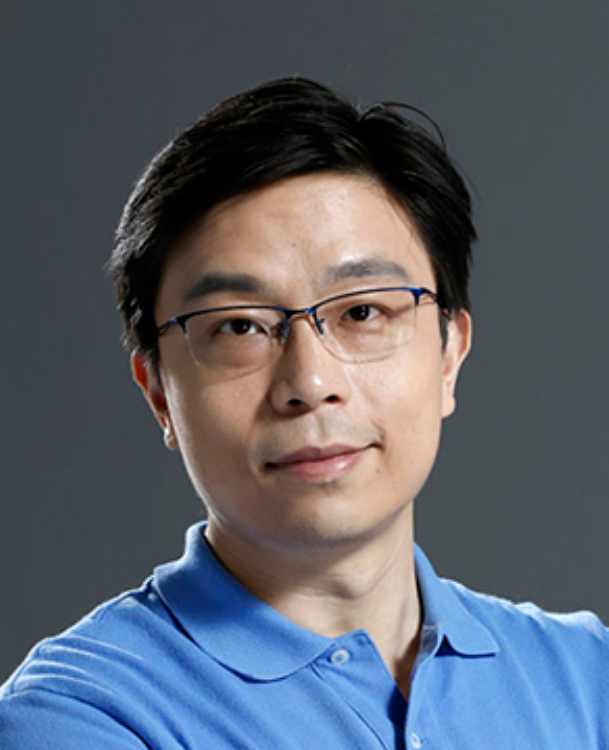}}]{Jiaya Jia} received the Ph.D.~degree in Computer Science from Hong Kong University of Science and Technology in 2004 and is currently a full professor in Department of Computer Science and Engineering at the Chinese University of Hong Kong (CUHK). He assumes the position of Associate Editor-in-Chief of IEEE Transactions on Pattern Analysis and Machine Intelligence (TPAMI) and is in the editorial board of International Journal of Computer Vision (IJCV). He continuously served as area chairs for ICCV, CVPR, AAAI, ECCV, and several other conferences for the organization. He was on program committees of major conferences in graphics and computational imaging, including ICCP, SIGGRAPH, and SIGGRAPH Asia. He is a Fellow of the IEEE. 
\end{IEEEbiography}